\documentclass[a4paper,fleqn]{cas-dc}
\let\printorcid\relax 

\usepackage[numbers]{natbib}
\def\tsc#1{\csdef{#1}{\textsc{\lowercase{#1}}\xspace}}
\tsc{WGM}
\tsc{QE}

\pdfoutput=1

\begin{document}
\let\WriteBookmarks\relax
\def\floatpagepagefraction{1}
\def\textpagefraction{.001}

\shorttitle{Facial Expression Analysis Using DMSNs}    

\shortauthors{W.C. de Melo~et al.}  

\title [mode = title]{Facial Expression Analysis Using Decomposed Multiscale Spatiotemporal Networks}



%

\author[1]{Wheidima Carneiro {de Melo}}[
                        ]
                        
\cormark[1]


\ead{wheidima.melo@oulu.fi}



\affiliation[1]{organization={Center for Machine Vision and Signal Analysis (CMVS), University of Oulu},
            country={Finland}}

\author[2]{ Eric Granger}[%
   ]
\ead{eric.granger@etsmtl.ca}

\affiliation[2]{organization={Laboratoire d’imagerie, de vision et d’intelligence artificielle (LIVIA), École de Technologie Supérieure},
            country={Canada}}

\author[1,3]{ Miguel Bordallo Lopez}[%
   ]
\ead{miguel.bordallo@vtt.fi}

\affiliation[3]{organization={VTT Technical Research Centre of Finland Ltd},
            country={Finland}}
            
\cortext[1]{Corresponding author}
\nonumnote{W.C. de Melo is also affiliated with Amazonas State University}

\begin{abstract}
Video-based analysis of facial expressions has been increasingly applied to infer health states of individuals, such as depression and pain. Among the existing approaches, deep learning models composed of structures for multiscale spatiotemporal
processing have shown strong potential for encoding facial dynamics. However, such models have high computational complexity, making for a difficult deployment of these solutions. To address this issue, we introduce a new technique to decompose the extraction of multiscale spatiotemporal features. Particularly, a building block structure called Decomposed Multiscale Spatiotemporal Network (DMSN) is presented
along with three variants: DMSN-A, DMSN-B, and DMSN-C blocks. The DMSN-A block generates multiscale representations by analyzing spatiotemporal features
at multiple temporal ranges, while the DMSN-B block analyzes spatiotemporal features at multiple ranges, and the DMSN-C block analyzes spatiotemporal features at multiple spatial sizes. Using these variants, we design our DMSN architecture which has the ability to explore a variety of multiscale spatiotemporal features, favoring the adaptation to different facial behaviors. Our extensive experiments on challenging datasets show that the DMSN-C block is effective for depression detection, whereas the DMSN-A block is efficient for pain estimation. Results also indicate that our DMSN architecture provides a cost-effective solution for expressions that range from fewer facial variations over time, as in depression detection, to greater variations, as in pain estimation.
\end{abstract}



\maketitle

\section{Introduction}\label{sec:intro}

Given the population growth, and global shortage of doctors, among others, healthcare applications have been driving the development of automatic systems for medical diagnosis. Such technology can be beneficial to improve the quality of clinical outcomes, and the access to healthcare services. Since face can provide information concerning medical conditions~\cite{mirror}, there has been a growing interest in developing contact-free, objective, and accurate systems for automatic assistive medical diagnosis from facial videos~\cite{mirror,surveyDep,surveyPain}. These video-based methods encode the correlations between appearance and dynamics of facial expressions and health states of an individual. For instance, Jaiswal~\textit{et al.}~\cite{autism} proposed a method that explores facial expressions, and head pose and movement to predict Attention Deficit Hyperactivity Disorder (ADHD), and Autism Spectrum Disorder (ASD). 


Two emerging applications for automatic facial expression analysis are depression detection and pain estimation. Depression is defined as a negative state of mind which remains for a long period of time. Such a mental health disorder can affect an individual's emotions, behavior, mind, and physical health~\cite{physical}. In severe conditions, depression conducts to substance abuse and suicidal behavior~\cite{suicidal}. Despite the existence of effective treatment, it is estimated that, in Europe, about $56\%$ of patients suffering from depression receive no treatment~\cite{europe}. The reasons for this high number include client fees, and restricted or lack of accessibility to mental healthcare. Studies also show that clinicians have difficulties to diagnose depression~\cite{diag1,diag2}. Indeed, the assessment of depression has a subjective nature since it relies on doctor's perception of patient reports. Inaccurate diagnosis of depression has produced an alarming number
of false-positives that present grave consequences for the patients~\cite{diag2}.

Pain is an important physical sign associated with the health conditions of an individual. It can be considered as a highly disturbing sensation caused by injury, illness or mental distress, and it is related to depression~\cite{painDep}. The clinical evaluation of pain is mainly determined by patient self-reports (e.g., by using Visual Analogue Scale (VAS)~\cite{vas} or Numeric Rating Scale (NRS)~\cite{nrs}). However, the assessment provided by a patient may not be reliable since patients may have restricted communication potential (e.g., neonates), cognitive impairments or are under the influence of medication. An alternative is the medical staff (e.g., doctors and nurses) perform the assessment. However, observers may overestimate or underestimate pain intensity which impair the treatment~\cite{painQuestion}, and the continuous monitoring is impracticable.

Automatic analysis of facial variations for objective recognition of expressions associated with health states like depression and pain can assist in the reliability and improvement of clinical assessment and monitoring, as well as mitigate issues regarding accessibility and costs. Studies have found facial cues related to depression, such as limited facial expressiveness~\cite{express}, reduced eye contact~\cite{rEye}, smiles with a shorter duration and less intensity~\cite{smile}, and a small number of mouth movements~\cite{mouth}. In contrast, facial expressions~\cite{pain1,surveyPain} involving, e.g., closed eyes, raised cheeks, and a wrinkled nose are relevant indicators of pain. With that, we can claim that a pain event may produce expressions with greater facial variations over time, and a depressive state is linked to expressions with fewer variations over time. Therefore, systems for facial expression analysis based on videos can explore these cues to predict depressive or painful states.

Recently, the emergence of state-of-the-art deep learning (DL) architectures has contributed to significant progress in diverse visual recognition tasks, such as action recognition~\cite{i3d}, image classification~\cite{imageNet}, and activity understanding~\cite{activity}. DL models have also been shown to provide a high level of predictive accuracy for automatic facial expression analysis from videos~\cite{surveyPain,lopez,msn,mdn,scn}. Given the availability of pre-trained models for still images, DL models commonly employ 2D Convolutional Neural Networks (CNNs) to leverage spatial correlations, along with an aggregation scheme or a recurrent technique to capture temporal dependencies~\cite{icip,attentionDep,FDHHDep,DTLDep,VGG16LstmPain,RcnnPain,PoolingDep,FourResNetDep}. Such an approach has limited capacity in encoding important dynamic information~\cite{msn,mdn,scn}. Conversely, 3D CNNs can directly model spatiotemporal variations in facial information from input video clips~\cite{fg,c3dDep,weaklyPain}. However, in addition to high computational complexity, these architectures use basic building blocks that explore a fixed spatiotemporal range, which limits the ability to learn discriminative features since facial expression variations comprise various ranges, and the difference of these variations along distinct levels of a health condition can be small.

In other application domains, efficient architectures have been developed for the modeling of spatiotemporal information~\cite{tsm,p3d,tsn,bottomTopI3D}. However, these methods also rely on structures with the ability to explore fixed
spatiotemporal range. To address this problem, some works~\cite{msn,mdn,scn} present effective architectures to model facial expression variations in videos. Such methods explore multiscale spatiotemporal features by using either parallel 3D convolutions with different kernels~\cite{msn,scn} or multiple structures that explore different spatiotemporal ranges~\cite{mdn}. Although these approaches achieve a high level of performance, the models have a high number of parameters and computations, even when compared to 3D CNNs.

\begin{figure}[!t]
\centering
\includegraphics[scale=0.47]{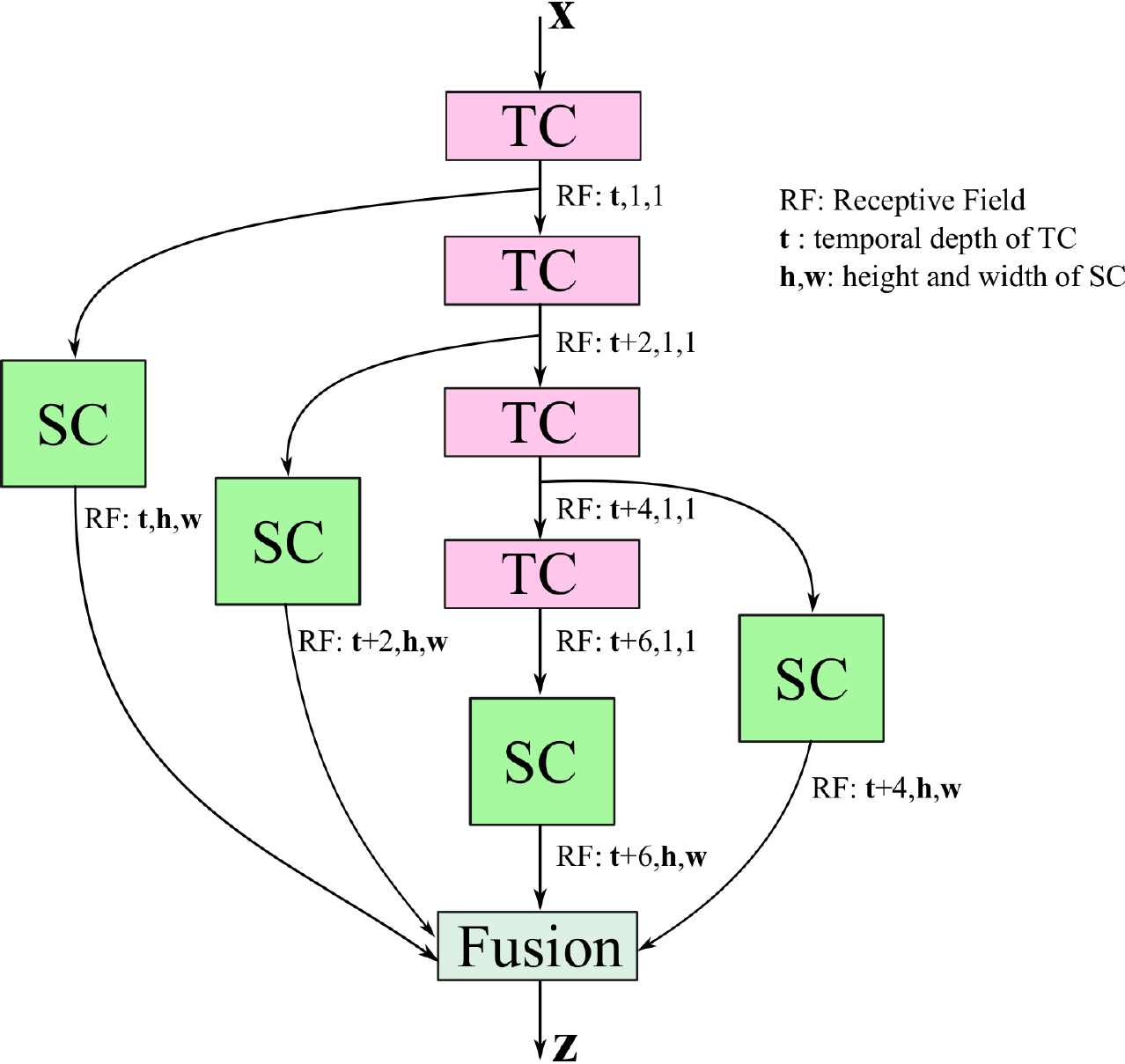}
\caption{A variant of proposed DMSN block. Each temporal convolution (pink block) generates features at different scales. Spatial convolutions (green block) complement this operation to explore spatiotemporal features. The fusion stage combines the multiscale spatial and temporal features. A detailed description of this block is presented in Section~\ref{sec:dmsn}.}
\label{figExample}
\end{figure}

In this paper, we propose an efficient alternative for the modeling of facial expression variations captured in videos. The method decomposes the exploration of multiscale spatiotemporal information to reduce computational costs. Specifically, we introduce a building block called Decomposed Multiscale Spatiotemporal Network (DMSN). The structure consists of a sequence of convolutions to produce multiscale features, where every element operates on a domain, and the branches of this sequence operate on a complementary domain of these elements, allowing to generate multiscale spatiotemporal representations. This design allows the development of three different blocks: DMSN-A, DMSN-B, and DMSN-C. The DMSN-A block learns spatiotemporal features
with distinct temporal ranges at a fixed spatial size (see Fig.~\ref{figExample}). The DMSN-B block explores diverse spatiotemporal features at distinct ranges. Lastly, the DMSN-C block analyzes spatiotemporal features with different spatial sizes
at a fixed temporal range. Our proposed blocks employ residual connections, and are implemented using only 1D and 2D convolutions. Using these three blocks, we design our DMSN architecture which has the potential to adapt to different facial behaviors thanks to the different multiscale spatiotemporal representation abilities of the proposed blocks.

The key contributions of this paper are as follows.
\begin{itemize}

\item A new building block structure is proposed with three variants -- DMSN-A, DMSN-B, and DMSN-C blocks -- to improve the extraction of multiscale spatiotemporal features. Such variants are employed in our DMSN architecture to provide discriminative representations for different facial behaviors.

\item We show empirically that our DMSN-C block is effective for exploring the spatiotemporal dependencies for depression detection whereas DMSN-A block is efficient to capture facial dynamics for pain estimation.

\item An extensive set of experiments on the challenging AVEC2013 and AVEC2014 depression datasets, and UNBC-McMaster and BioVid pain datasets, allowing to validate that our DMSN architecture can provide a level of performance that is comparable to state-of-the-art DL models, while significantly reducing
the computational costs.


\item An analysis of depression and pain features showing that depression features are more useful for pain estimation than pain features are for depression detection.
\end{itemize}

The rest of this paper is organized as follows. Section~\ref{sec:work} presents some background on methods for depression detection, pain estimation, and spatiotemporal modeling. Our DMSN architecture is introduced in Section~\ref{sec:dmsn}. Then, Section~\ref{sec:met} describes the experimental methodology, and Section~\ref{sec:results} discusses the results for validation of our approach. Finally, Section~\ref{sec:con} draws the conclusions of the present work.
 
\section{Related Work}\label{sec:work}
The growing interest in analyzing facial expressions captured in videos can be attributed to the psychological studies that indicate the correlation of a health condition and face, and the recent progress in deep learning and computer vision methods. The existing works try to explore non-verbal facial cues in order to infer health conditions. A key challenge is to obtain a robust representation in a scenario of subjective variability of facial expressions across different individuals and capture conditions.

\subsection{Models for depression detection}
Various authors proposed hand-engineered representations for depression detection. Some examples are the method proposed by Cohn~\textit{et al.}~\cite{cohn}, that employs Active Appearance Model (AAM) features and uses Support Vector Machine (SVM) as classifier, and the one proposed by Gupta~\textit{et al.}~\cite{lbpDep}, that uses Local Binary Pattern (LBP) features and Support Vector
Regressor (SVR). DL models have demonstrated more potential to extract discriminant features from spatiotemporal expressions correlated with depressive states. A common  approach is to employ a 2D CNN and some aggregation technique to explore facial features that are extracted from videos~\cite{icip,attentionDep,FDHHDep,DTLDep,PoolingDep,FourResNetDep}. For instance, Zhou~\textit{et al.}~\cite{PoolingDep} used ResNet-50 to explore the appearance information, and attention mechanism to fuse the static facial features. However, such methods have limited ability for the encoding of rich spatiotemporal variations in faces. Two-stream networks~\cite{icassp,twoStreamDep} and 3D CNNs~\cite{fg,c3dDep} have also been presented for depression detection. However, these methods are composed of structures that analyze a fixed spatiotemporal range, which reduces the ability to produce discriminative features. Indeed, it has been shown that a better multiscale capacity is favorable for depression detection which is characterized by small facial expression variations along
different levels~\cite{msn,mdn,song}. In this context, Song~\textit{et al.}~\cite{song} used spectral representations of behavior signals to analyze multiscale depression patterns. Two recent state-of-the-art methods--Multiscale Spatiotemporal Network (MSN)~\cite{msn}, and Maximization and Differentiation Network (MDN)~\cite{mdn}--have shown effectiveness in modeling multiscale spatiotemporal information. The structure of MSN is composed of 3D convolutions with different kernel sizes, whereas the one of MDN is formed using multiple maximization and difference blocks which explore features in diverse ranges. Although these methods achieve a high level of performance, their computational costs are expensive.

\subsection{Models for pain estimation}
Early methods for pain intensity estimation employed hand-engineered features such as LBP~\cite{lbpPain1}, Gabor~\cite{gaborPain}, Local Binary Patterns from Three Orthogonal Planes (LBP-TOP)~\cite{lbpTopPain}, Histograms of Topographical (HoT)~\cite{hotPain}, Pyramid Histogram of Orientation Gradients (PHOG)~\cite{phogPain} and Pyramid Local Binary Pattern (PLBP)~\cite{phogPain}. In recent years, DL models have been used to encode facial expression variations for pain estimation. Some methods generate deep representations by using frame-wise feature extraction~\cite{VGG16LstmPain,Vgg11LstmPain,RcnnPain}. For instance, Rodriguez~\textit{et al.}~\cite{VGG16LstmPain} employed VGG-16 architecture to learn spatial features and Long-Short Term Memory (LSTM) for the capturing of temporal relationships. Other works proposed to model spatiotemporal information within video sequences by employing 3D CNNs~\cite{weaklyPain,c3dPain}. Using this approach, Wang~\textit{et al.}~\cite{c3dPain} applied Convolutional 3D (C3D) network, that has as basic structure one $3\times3\times3$ convolutional layer, to recognize pain expressions. However, these two approaches are frequently ineffective to capture extensive range of facial expression variations. In~\cite{scn}, the authors presented evidence that a multiscale approach is more effective for the modeling of spatial and temporal dependencies related to pain status. They introduced the Spatiotemporal Convolutional Network (SCN) which employs as basic structure parallel 3D convolutions with different temporal depths. SCN obtains high performance for pain estimation, but it requires more than $500$M trainable parameters, making its deployment costly. 

\subsection{Spatiotemporal networks}
Since 3D CNNs have an ability to directly model spatial and temporal information, these methods are an intuitive choice for video analysis. Tran~\textit{et al.}~\cite{c3d} proposed an architecture with $8$ convolutional layers, called C3D, to learn spatiotemporal features. Carreira~\textit{et al.}~\cite{i3d} proposed to inflate all the filters and pooling kernels of 2D Inception model into 3D CNN generating Inflated 3D-ConvNet (I3D) model. Hara~\textit{et al.}~\cite{3dResnet} proposed 3D CNNs based on residual connections called 3D ResNet. In~\cite{slowFast}, the authors introduced the SlowFast network which consists of a slow path to model spatial semantics and a fast path to capture motion at fine temporal resolution. The principal drawbacks of employing 3D CNNs are the high computational complexity, and the lack of pre-trained backbone models. Recently, diverse architectures have been developed for efficient spatiotemporal modeling. In~\cite{tsn}, Temporal Segment Network (TSN) is introduced to model long-term temporal information employing 2D CNNs. Qiu~\textit{et al.}~\cite{p3d} proposed Pseudo-3D residual network (P3D) which factorizes 3D convolution into 2D and 1D convolution. Xie~\textit{et al.} studied I3D, I2D, as well as the combination of 2D and 3D methods by using either bottom-heavy (lower layers use 3D convolutions and upper layers use 2D convolutions) or top-heavy (lower layers use 2D convolutions and upper layers use 3D convolutions) networks. Lin~\textit{et al.}~\cite{tsm} proposed the Temporal Shift Module (TSM) to enable 2D CNNs to explore spatiotemporal dependencies by shifting channels along the temporal dimension. Even though such architectures are efficient for tasks like action recognition, their structure explores a fixed spatiotemporal range (e.g., P3D block uses a combination of one $1\times3\times3$ convolutional layer with one $3\times1\times1$ convolutional layer, encompassing a spatiotemporal receptive field size of $3\times3\times3$), which hinders the capacity of extracting effective representations for facial expression variations. Instead of exploring a fixed range, our proposed blocks explore multiple spatiotemporal ranges favoring the generation of discriminative representations.

In contrast to existing methods that use a structure to explore multiscale spatiotemporal information (i.e., MSN, MDN, and SCN), our proposed blocks are designed to efficiently capture such information for the representation of facial videos. To achieve this goal, three distinct variants of the DMSN block are proposed to decompose the extraction of multiscale spatiotemporal features. The use of these three variants allows the design of a cost-effective DMSN architecture. Another unique benefit of our architecture is the ability to adapt to distinct facial behaviors,  since it is built employing our three proposed blocks.

\begin{figure}[!t]
\centering
\includegraphics[scale=0.245]{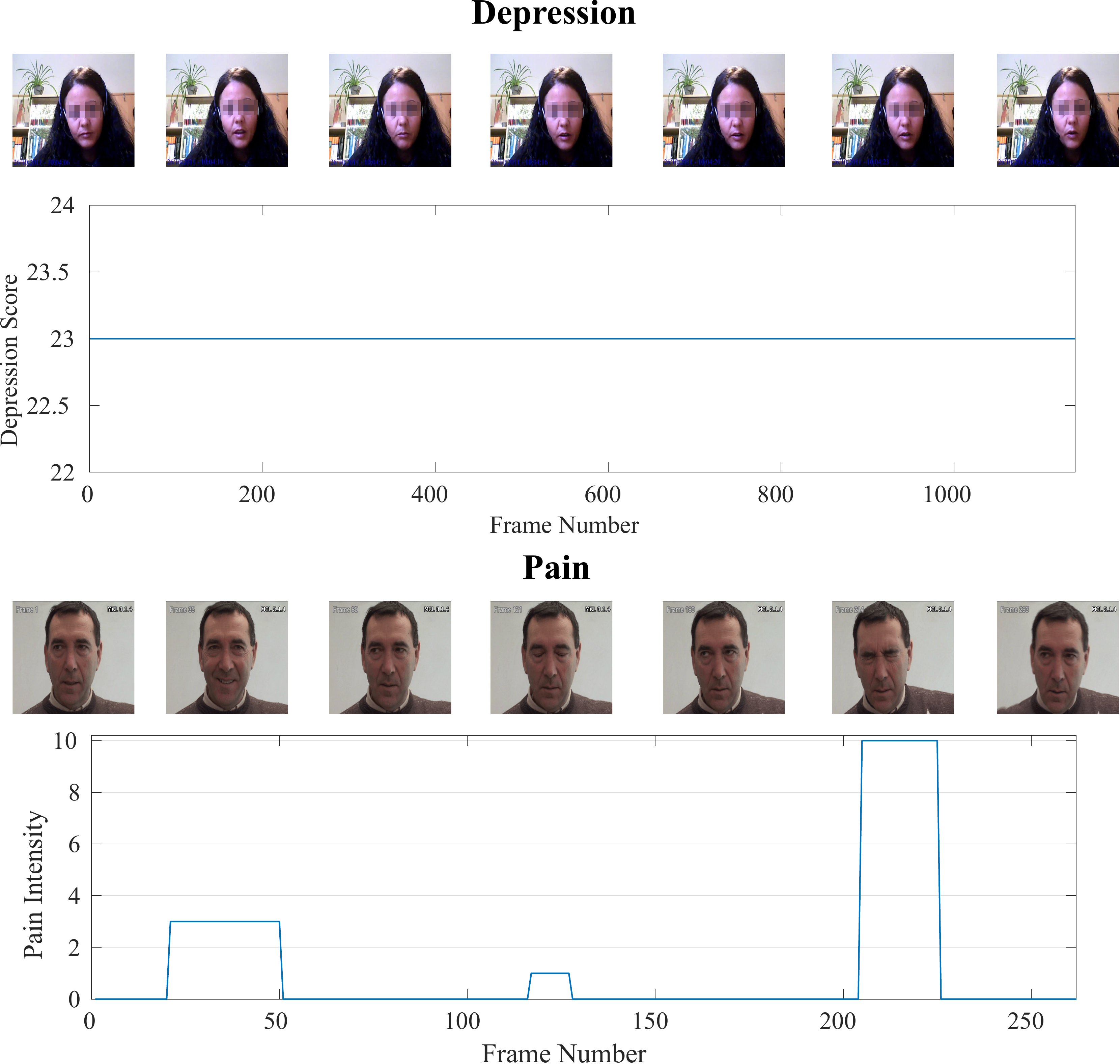}
\caption{Examples of depression and pain expressions in sequences of consecutive video frames. The depression level is constant over the video sequence, whereas pain level may change rapidly over time.}
\label{fig1}
\end{figure}

\section{Decomposed Multiscale Spatiotemporal Network} \label{sec:dmsn}

The dynamics of facial expressions provide rich information for the recognition of facial patterns related to a health condition. Such facial expression variations can be explored, e.g., velocity or intensity, in order to model different levels of a health state. This work aims to develop a deep architecture to capture an extensive range of facial dynamics to produce efficient representations for automatic facial expression analysis. Specifically, we design the Decomposed Multiscale Spatiotemporal Network (DMSN) by introducing three multiscale convolution blocks that employ different strategies to generate multiscale spatiotemporal representations.

We design our DMSN blocks considering that the facial behavior can considerably differ in two distinct health diagnosis applications. As illustrated in Fig.~\ref{fig1}, the level of pain can change over time, and the correlated facial expressions can be modified considerably over a short period. On the other hand, the depression level lasts for a longer period and the resulting facial expressions tend to have more gradual variations. Consequently, an effective architecture for facial expression analysis has the capability of adaptation to distinct facial behaviors. This fact motives us to build our architecture using blocks with different abilities.

\begin{figure*}[htb]
\centering
\includegraphics[scale=0.33]{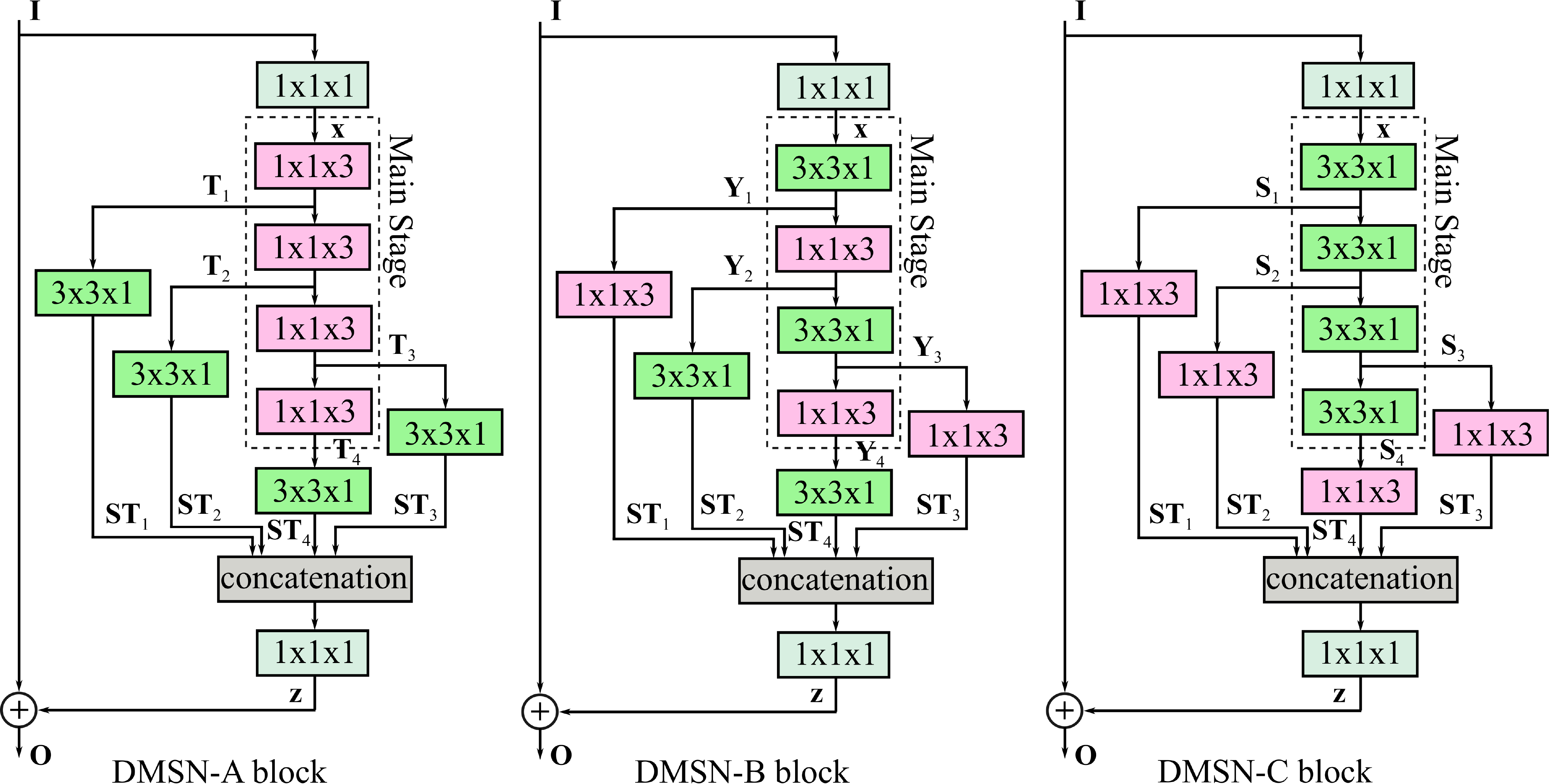}
\caption{The three proposed DMSN building blocks. Pink blocks represent temporal convolutions whereas green blocks represent spatial convolutions. Each convolution in the Main Stage extends the range of features, and its branches use a complementary filter to explore spatiotemporal features. Features of these branches are at different scales, and are combined using a $1\times1\times1$ convolution.}
\label{fig2}
\end{figure*}

To develop our proposed blocks, we employ a sequence of convolutions to increase the range of the region under  analysis. This sequence is called Main Stage sub-block (see Fig.~\ref{fig2}). The output of each convolution in the Main Stage is connected as the input to another convolution that operates in a complementary domain to encode spatiotemporal information. The output feature maps of these branches are at different scales, and a $1\times1\times1$ convolution fuses these features to generate multiscale spatiotemporal representations.

The architectural design of our DMSN block allows the investigation of different strategies to extract multiscale spatiotemporal features. Since the Main Stage sub-block is responsible for the multiscale ability, it is able to employ
convolutions on either the same or different domains, which can be beneficial in the elaboration of more efficient multiscale representations. In this context, we derive three variants of our proposed block (see Fig.~\ref{fig2}). In the sequence, we present a detailed description of these variants.

\subsection{DMSN-A block}
Considering that the pain level can vary more rapidly over time, its level can last for different periods, and it can produce sudden facial expression variations, we define the Main Stage of the DMSN-A block as a sequence of $3\times1\times1$ temporal convolutions. This sub-block is formed using four 1D temporal convolutions in order to explore the short, medium, and long temporal ranges. The output $\mathbf T_i$ of each 1D temporal convolution ($\mathbf M^t_i$) is given by: 
\begin{equation}
\mathbf{T}_i = \begin{cases} \mathbf{M}^t_i(\mathbf{x}) & i=1 \\ \mathbf{M}^t_i(\mathbf{T}_{i-1}) & 2 \leq i \leq 4 \end{cases}
\end{equation}
\noindent Each 1D convolution increases the temporal range explored by this sub-block. Branches of the Main Stage employ $1\times3\times3$ spatial convolutions which generate spatiotemporal features at multiple temporal ranges. The output $\mathbf{ST}_j$ of each 2D spatial convolution ($\mathbf M^s_j$) is defined by: 
\begin{equation}
\mathbf{ST}_j = \mathbf{M}^s_j(\mathbf{T}_{j}) \quad 1 \leq j \leq 4
\end{equation} 

\subsection{DMSN-B block}
This block employs the Main Stage sub-block to increase the explored regions in both domains by using $1\times3\times3$ spatial convolutions and $3\times1\times1$ temporal convolutions. With the purpose of maintaining a similar computational complexity in comparison with DMSN-A block, the DMSN-B block employs four convolutions in the Main Stage. The output $\mathbf Y_i$ of each element in this sub-block is calculated by:
\begin{equation}
\mathbf{Y}_i = \begin{cases} \mathbf{M}^s_i(\mathbf{x}) & i=1 \\ \mathbf{M}^s_{i-1}(\mathbf{Y}_{i-1}) & i=3 \\ \mathbf{M}^t_{i/2}(\mathbf{Y}_{i-1}) & i=2,4\end{cases}
\end{equation}
\noindent Each element of this sub-block increases the spatiotemporal receptive field size in analysis. Branches of the Main Stage use complementary convolution (in relation to domain) to generate spatiotemporal features at multiple ranges. Specifically, the output $\mathbf{ST}_j$ of each branch is given by:
\begin{equation}
\mathbf{ST}_j = \begin{cases} \mathbf{M}^t_{j+3-(j+1)/2}(\mathbf{Y}_{j}) & j=1,3 \\ \mathbf{M}^s_{j+2-j/2}(\mathbf{Y}_{j}) & j=2,4\end{cases}
\end{equation} \newline

\subsection{DMSN-C block}
Given that depressive states can present less facial expression variations over time, and the depression level of a subject in a video tends to be constant, the DMSN-C block employs the Main Stage to produce multiscale spatial features. The sub-block is constituted by a sequence of $1\times3\times3$ spatial convolutions where each element increases the spatial receptive field size. The output $\mathbf S_i$ of each 2D spatial convolution ($\mathbf M_i^s$) is defined by: 
\begin{equation}
\mathbf{S}_i = \begin{cases} \mathbf{M}^s_i(\mathbf{x}) & i=1 \\ \mathbf{M}^s_i(\mathbf{S}_{i-1}) & 2 \leq i \leq 4 \end{cases}
\end{equation}
\noindent Branches of the Main Stage use $3\times1\times1$ temporal convolution to produce spatiotemporal features at multiple spatial sizes. The output $\mathbf{ST}_j$ of each 1D temporal convolution ($\mathbf M^t_j$) can be given by
\begin{equation}
\mathbf{ST}_j = \mathbf{M}^t_j(\mathbf{S}_{j}) \quad 1 \leq j \leq 4
\end{equation}

Furthermore, for DMSN-A, DMSN-B, and DMSN-C blocks, the first element of the Main Stage reduces the number of channels by half in comparison with the number of output channels of the first $1\times1\times1$ convolution whereas the convolutions in the branches reduces this number by one quarter (i.e., 1 divided by the number of branches).

\subsection{DMSN architecture}
We construct the DSMN architecture using our three blocks. In this way, our model has structures with diverse capacities favoring the creation of a model that can perform well in different applications. In Table~\ref{tab1}, we provide the details of our proposed model. The output feature map is defined as a tensor $\mathbf X \in \mathbb{R}^{T\times H \times W \times C}$, where $T$, $H$, $W$, and $C$ are the temporal depth, height, width, and number of channels, respectively. The model size and the number of blocks in each layer are defined similarly to ResNet-50. The DMSN blocks are employed in the residual layers (res) and the regression layer outputs a value related to pain or depression score. Moreover, we develop three models which are named according to the DMSN block they employ, e.g.,
DMSN-A model uses only DMSN-A blocks. With that, we can understand the contributions of each DMSN block for a given application. 

\begin{table*}[htb]
\centering
\caption{\label{tab1} Properties of the proposed DMSN architecture. DMSN-A, DMSN-B, and DMSN-C models are built using only an instance of DMSN block. For example, DMSN-A only uses the DMSN-A block.}
{
\begin{tabular}{c|c|c|c|c}
\hline
\textbf{Layer} & \textbf{Output Size} & \textbf{Number of Channels} &\textbf{Structure} & \textbf{Number of Layers} \\
\hline \hline
input & $16\times112\times112$ &$3$&  & $\times1$\\
\hline
conv1 & $16\times56\times56$&$64$& $7\times7\times7$ & $\times1$\\
\hline
MaxPool & $8\times28\times28$&$64$& $3\times3\times3$ & $\times1$\\
\hline
\multirow{3}{*}{res2} & \multirow{3}{*}{$8\times28\times28$}&\multirow{3}{*}{128}& DMSN-A & \multirow{3}{*}{$\times1$}\\
&&&DMSN-B&\\
&&&DMSN-C&\\
\hline
\multirow{4}{*}{res3} & \multirow{4}{*}{$8\times14\times14$}&\multirow{4}{*}{256}& DMSN-A & \multirow{4}{*}{$\times1$}\\
&&&DMSN-B&\\
&&&DMSN-C&\\
&&&DMSN-A&\\
\hline
\multirow{3}{*}{res4} & \multirow{3}{*}{$8\times7\times7$}&\multirow{3}{*}{512}& DMSN-A & \multirow{3}{*}{$\times2$}\\
&&&DMSN-B&\\
&&&DMSN-C&\\
\hline
\multirow{4}{*}{res5} & \multirow{4}{*}{$8\times4\times4$}&\multirow{4}{*}{1024}& DMSN-A & \multirow{4}{*}{$\times1$}\\
&&&DMSN-B&\\
&&&DMSN-C&\\
&&&DMSN-A&\\
\hline
regression & $1\times1$ & \multicolumn{2}{c}{spatial AvgPool, FC, AvgPool}\\

\hline
\end{tabular}
}
\end{table*}

\section{Experimental Methodology} \label{sec:met}

\subsection{Depression datasets}
We conduct experiments on two public benchmarking datasets for depression, called the Audio-Visual Emotion Challenge 2013 and 2014 (AVEC2013 \cite{avec2013} and AVEC2014 \cite{avec2014} depression sub-challenge datasets). This AVEC sub-challenge consisted of predicting the depression severity of subjects on Beck Depression Inventory (BDI-II). The severity of depression can be determined in accordance to BDI-II score as follows: minimal ($0-13$), mild ($14-19$), moderate ($20-28$), and severe ($29-63$). Although there exist other depression datasets such as AVEC2016~\cite{avec2016}, to the best of our knowledge, AVEC2013 and AVEC2014 are the only datasets that provide raw video data.

The AVEC2013 dataset is composed of $150$ videos from a group of individuals which the average age is $31.5$ years. The individuals were recorded during an interaction with a computer carrying out $14$ tasks, including counting from $1$ to $10$. The dataset is divided into three partitions: training, development, and test subsets. Every subset is comprised of $50$ videos, where each video has a BDI-II score as a discrete-value label which indicates the level of depression of an individual. The maximum duration of videos is $50$ minutes, the minimum is $20$ minutes, and the average length is $25$ minutes.

\begin{figure}[htb]
\centering
\includegraphics[scale=0.17]{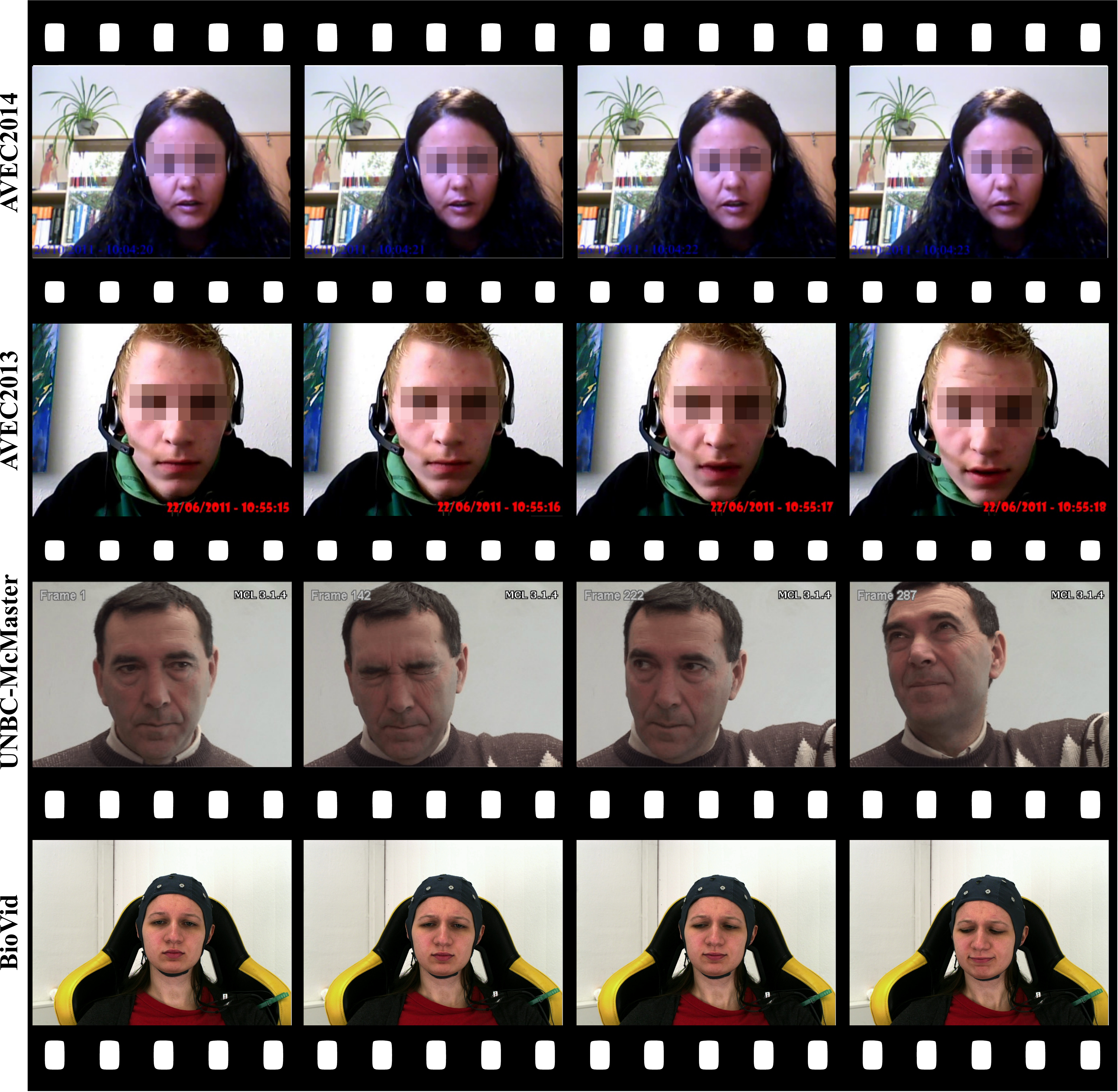}
\caption{Examples of facial frames from AVEC2014 and AVEC2013 depression datasets as well as UNBC-McMaster and BioVid pain datasets.}
\label{figData}
\end{figure}

The AVEC2014 dataset contains videos of individuals performing two tasks: Freeform and Northwind. In the first one, the individuals answer questions like discussing a sad childhood memory. In the second one, individuals read audibly an excerpt from a fable. In total, there are $150$ videos of each task with a ground truth label (BDI-II score) for each video. For both tasks, the videos are distributed in three partitions: training, development, and test subsets. The videos have length between $6$ and $248$ seconds. Samples from both datasets are exhibited in Fig.~\ref{figData}. Due to privacy concerns, all samples of depression shown in this work are blurred.

\subsection{Pain datasets}
To evaluate the performance of our proposed approach on pain estimation, we
conduct experiments on two publicly available datasets: UNBC-McMaster Shoulder Pain Expression Archive~\cite{unbc}, and BioVid Heat Pain~\cite{biovid}.

The UNBC-McMaster dataset has been largely employed for pain estimation from facial information. It consists of $200$ face videos of $25$ individuals with a total number of $48,398$ frames. Fig.~\ref{figData} presents some facial frames from this dataset. Each video is labeled using Prkachin and
Solomon Pain Intensity (PSPI) scores in a frame-level fashion on a range of $16$ discrete levels ranging from $0$ (no pain) to $15$ (maximum pain). Since the input of our proposed model is a clip, we follow the works in~\cite{mdn,scn,weaklyPain,daPain,labelPain,selfPain}, which define a label for each clip. Specifically, we use the average of the pain intensity of each frame inside the clip as a label. Moreover, since the dataset is highly imbalanced ($82.7\%$ of frames have pain score of $0$), we adopted the common quantization strategy, which maps the pain levels to $6$ ordinal levels as: $0$:$0$, $1$:$1$, $2$:$2$, $3$:$3$, $4$-$5$:$4$, $6$-$15$:$5$.

The BioVid Heat Pain dataset contains videos and bio-signals that were acquired during acute heat-induced pain experiments in healthy adults. Pain was induced in four distinct intensities in the right arm of each individual. Although the dataset includes bio-signals such as Skin Conductance Level (SCL), electrocardiogram (ECG), electromyography (EMG), and electroencephalogram (EEG), our experiments only consider Biovid part A which has $8,700$
videos of $87$ individuals. Each video is labeled with a pain stimulus level which ranges from $0$ (no pain) to $4$ (severe pain). A sample from this dataset is shown in Fig.~\ref{figData}.

\subsection{Training of the model}
 The model analyzes faces that are detected and extracted from video frames of datasets employing MTCNN~\cite{mtcnn}. Each facial image is resized to form a bounding-box sample with the size of $112\times 112\times 3$ that is fed to the model. Usually, datasets for facial expression analysis have a limited amount of training data, which can hinder the generalization ability of a deep architecture. To avoid this problem, deep models are normally pre-trained on large datasets and then fine-tuned on the target dataset. Following the works in~\cite{msn,mdn}, our proposed model is pre-trained on the VGGFace2 dataset~\cite{vggFace} that contains $3.31$ million images of more than $9,000$ subjects. In this process,
the model is optimized using Stochastic Gradient Descent (SGD) with a momentum of $0.9$, weight decay $0.0001$, and an initial learning rate of $0.01$. The learning rate is divided by $10$ after every $10$ epochs. The RGB input images are normalized by using the mean channel subtraction. In the fine-tuning process, the ADAM optimization algorithm is adopted. For depression detection task, the initial learning rate is defined as $0.005$, then, in the second epoch, this rate is modified to $0.0005$. The training is stopped after $3$ epochs. For pain estimation task, we define the learning rate equal to
$0.001$ under two epochs training. In the data augmentation process, we follow the same strategy as in~\cite{mdn,msn}.

\subsection{Performance measures}
For depression detection, an input video from the test subset is segmented into non-overlapped clips of $16$ frames. The model generates a depression score for each clip and the median of these values defines the final predicted score for the input video. In order to provide a fair comparison with state-of-the-art methods, we report the performance of the proposed architecture in terms of Mean Absolute Error (MAE) and Root Mean Square Error (RMSE), which are commonly used for depression detection~\cite{FourResNetDep,icip,attentionDep,mdn,c3dDep,fg}.
For pain estimation, we perform leave-one-subject-out cross-validation to evaluate the performance of our proposed model. For fair comparison with state-of-the-art methods, the performance of our architecture is measured in terms of Mean Square Error (MSE), and MAE, which are widely used for pain estimation~\cite{scn,mdn,VGG16LstmPain,Vgg11LstmPain}.
The computational complexity of models is assessed in terms of the number of parameters (memory complexity), and the number of Floating Point Operations (FLOPs) for the processing of a clip (time complexity).

\begin{table*}[htb]
\centering

\caption{\label{tab2} Performance of the proposed methods against spatiotemporal models for estimating of depression scores on AVEC2013 and AVEC2014 datasets.}
\label{tabDep}
{
\begin{tabular}{l||cc|cc|c|c}
\hline
\multirow{2}{*}{\bf{Architecture}} & \multicolumn{2}{c|}{\textbf{AVEC2013}} & \multicolumn{2}{c|}{\textbf{AVEC2014}} & \multirow{2}{*}{\bf{Parameters $\downarrow$}} & \multirow{2}{*}{\bf{FLOPs $\downarrow$}} \\\cline{2-5}
& \bf{RMSE} & \bf{MAE} & \bf{RMSE} & \bf{MAE} & \\
\hline
\hline
3D-ResNet~\cite{3dResnet} & 8.81 & 6.92 & 8.40 & 6.79 & 63.0M & 12.22G\\
TSN~\cite{tsn}       & 8.89 & 6.21 & 8.72 & 6.45 & 23.5M & 16.45G\\
TSM~\cite{tsm}       & 8.89 & 6.41 & 8.53 & 6.29 & 23.5M & 16.45G\\
P3D~\cite{p3d}       & 8.50 & 6.24 & 8.63 & 6.80 & 24.9M & 8.56G\\
\hline
DMSN-A (Ours)  & 7.98 & 6.32 & 8.13 & 6.48 & 19.0M & 10.26G\\
DMSN-B (Ours) & 7.92 & 6.59 & 7.86 & 6.24 & 23.6M & 10.83G\\ 
DMSN-C (Ours) & 7.77 & \bf{6.14} & 7.66 & 6.10 & 25.9M& 11.53G\\
DMSN (Ours) & \bf{7.66} & \bf{6.14} & \bf{7.50} & \bf{5.69} & 22.1M& 11.29G\\
\hline
\end{tabular}
}
\end{table*}

\section{Results and Discussion} \label{sec:results}

\subsection{Analysis of the DMSN blocks}
In order to investigate the potential of the proposed DMSN blocks, we generate results for the three models that are named according to the DMSN block they employ, e.g., DMSN-A uses DMSN-A blocks. We also compare these models with our proposed DMSN architecture to show the benefits of using all DMSN blocks. Finally, we compare our architecture in terms of performance and computational complexity with 3D ResNet~\cite{3dResnet} and three other efficient spatiotemporal models: TSN~\cite{tsn}, TSM~\cite{tsm}, and P3D~\cite{p3d}. For fair comparison, all these models follow the same training process that our proposed architecture, i.e., first pre-train on VGGFace2 dataset, then fine-tune on depression or pain datasets.

\subsubsection{Depression detection}
Table~\ref{tab2} reports the results for our three models on AVEC2013 and AVEC2014 datasets. When compared with DMSN-A, DMSN-B achieves better performance, except for AVEC2013 in terms of MAE. As can be seen, the best performance is obtained by DMSN-C. Regarding the computational complexity, it is possible to observe that DMSN-A employs fewer parameters and requires fewer FLOPs, whereas DMSN-C is more computationally expensive in comparison with DMSN-A, and DMSN-B. Among our three models, DMSN-C provides the best trade-off between performance and computational complexity since this model improves the results with slightly more resources. From these results, we can claim that the DMSN-C block is effective to explore facial expression variations for depression detection.

Table~\ref{tab2} also shows the performance of our proposed DMSN architecture which employs DMSN-A, DMSN-B, and DMSN-C blocks. The use of our three blocks in our architecture provides an improvement of results over DMSN-A, DMSN-B, and DMSN-C models (except for AVEC2013 in terms of MAE where DMSN-C achieves the same result). Observe that DMSN architecture has lower computational costs than the DMSN-C model. Although our architecture has higher FLOPs than DMSN-B and is more expensive than DMSN-A, DMSN significantly improves the performance on depression detection when compared with these two models. These results demonstrate that the diversity of multiscale spatiotemporal features explored by our DMSN architecture enhances the representation for recognition of depressive states.





We also compare our DMSN architecture with the 3D ResNet, TSN, TSM, and P3D models in Table~\ref{tab2}. DMSN improves the results by more than $1.0$ in terms of RMSE on AVEC2013 and in terms of MAE on AVEC2014 when compared with 3D ResNet. DMSN also outperforms TSN, TSM, and P3D where the difference in results on AVEC2014 is significant. DMSN employs fewer parameters than these models and has fewer FLOPs, except for P3D. As indicated by the results, DMSN has the potential to generate efficient spatiotemporal representations for depression detection.

\begin{table}[h]
\centering
{

\caption{\label{tab3} Performance of our proposed approach against spatiotemporal models on UNBC-McMaster and BioVid datasets.}

\begin{tabular}{l||cc|cc}
\hline
\multirow{2}{*}{\bf{Architecture}} & \multicolumn{2}{c|}{\textbf{UNBC-McMaster}} & \multicolumn{2}{c}{\textbf{BioVid}}\\ \cline{2-5}
& \bf{MSE} & \bf{MAE} & \bf{MSE} & \bf{MAE}\\
\hline
\hline
3D-ResNet~\cite{3dResnet} & 0.75 & 0.56 & 2.28 & 1.30 \\
TSN~\cite{tsn} & 0.58 & 0.53 & 2.07 & 1.21 \\
TSM~\cite{tsm} & 0.46 & 0.49 & 1.94 & 1.20 \\
P3D~\cite{p3d} & 0.67 & 0.50 & 2.04 & 1.23 \\
\hline
DMSN-A (Ours) & 0.43 & 0.39 & 1.68 & 1.08  \\
DMSN-B (Ours) & 0.41 & 0.37 & 1.70 & 1.09  \\ 
DMSN-C (Ours) & 0.44 & 0.38 & 1.71 & 1.09   \\
DMSN (Ours) & \bf{0.38} & \bf{0.35} & \bf{1.54} & \bf{1.04}\\

\hline
\end{tabular}

}
\end{table}

\subsubsection{Pain Estimation}
Table~\ref{tab3} presents the performance of our DMSN-A, DMSN-B, and DMSN-C models on UNBC-McMaster and BioVid datasets. As can be seen, the three DMSN models achieve comparable results on UNBC-McMaster pain dataset. On the other hand, DMSN-A exhibits a better performance on BioVid pain dataset when compared to DMSN-B and DMSN-C. Given that the DMSN-A model requires fewer parameters and FLOPs, the DMSN-A block, which has the capacity to explore diverse spatiotemporal features at different temporal ranges, can be considered an efficient strategy to capture spatiotemporal variations for pain estimation.

\begin{table}[htb]
\centering
{

\caption{\label{tab4} Performance analysis of our proposed DMSN architecture for different input depths on AVEC2014 and UNBC-McMaster datasets.}
\begin{tabular}{c||cc|cc|c}
\hline
\multirow{2}{*}{\bf{Depth}} & \multicolumn{2}{c|}{\textbf{AVEC2014}}  &  \multicolumn{2}{c|}{\textbf{UNBC-McMaster}} & \multirow{2}{*}{\bf{FLOPs$\downarrow$}}\\\cline{2-5}
& \bf{RMSE} & \bf{MAE} & \bf{MSE} & \bf{MAE} &\\
\hline
\hline
8 & 8.84 & 6.71 & 0.49 & 0.40 & $5.64$G\\
16 & \bf{7.50} & \bf{5.69} & \bf{0.38} & \bf{0.35} & $11.29$G\\
24 & \bf{7.50} & 5.80 & 0.43 & 0.36 & $16.93$G\\
32 & 7.72 & 5.96 & 0.38 & 0.40 & $22.57$G\\
\hline
\end{tabular}

}
\end{table}

In Table~\ref{tab3}, we also show the results of our DMSN architecture for pain estimation. The employment of the three DMSN blocks in our architecture produces better results in comparison with DMSN-A, DMSN-B, and DMSN-C models. Consequently, the construction of our architecture using different strategies to learn multiscale spatiotemporal features favors a performance improvement for an application with greater facial expression variations as in pain estimation, and one with fewer facial variations as in depression detection.

A comparison between our DMSN architecture and 3D ResNet, TSN, TSM, and P3D models is also presented in Table~\ref{tab3}. Compared with P3D, DMSN improves the results by $0.5$ and $0.29$ in terms of MSE on BioVid and UNBC-McMaster datasets, respectively. In summary, DMSN outperforms these methods and the difference in results is higher on BioVid dataset, indicating that DMSN has good ability to encode facial dynamics for pain estimation.

\subsection{Analysis of temporal depth of input}
Our DMSN architecture is designed to explore a wide range of facial expression variations. Consequently, the temporal depth of input is an important factor in the performance of the model. In Table~\ref{tab4}, we perform evaluations considering inputs with $8$, $16$, $24$, $32$ frames. Since the pain and depression datasets are composed of similar face videos, and the evaluations involve a long training process, we carry out this analysis on AVEC2014 and UNBC-McMaster datasets. For depression detection, using sequences with $8$ frames significantly degrades the performance of our model. In fact, very short sequences increase the level of ambiguity along the depression levels, making harder to generate effective representations. The model sustains the highest levels of performance for a clip size of $16$ and $24$ frames, and worsens for $32$ frames. For pain estimation, the model maintains a comparable level of performance for all sequences employed, but the worst results are obtained using clips with $8$ frames. Furthermore, as the clip size increases, the model requires more FLOPs to generate an output.

\begin{table*}[!h]
\centering
{
\caption{\label{tab5} Evaluation of our DMSN architecture considering different number of branches in the Main Stage sub-block on AVEC2014 and UNBC-McMaster datasets.}
\begin{tabular}{c||cc|cc|c|c}
\hline
\multirow{2}{*}{\bf{Number of branches}} & \multicolumn{2}{c|}{\textbf{AVEC2014}} & \multicolumn{2}{c|}{\textbf{UNBC-McMaster}} & \multirow{2}{*}{\bf{Parameters$\downarrow$}} & \multirow{2}{*}{\bf{FLOPs$\downarrow$}} \\ \cline{2-5}
& \bf{RMSE} & \bf{MAE} & \bf{MSE} & \bf{MAE} & \\ \cline{2-5}
\hline
\hline
2 & 8.45 & 6.64 & 0.63 & 0.52 & 18.0M & 9.64G\\
3 & 7.71 & 6.08 & 0.45 & 0.42 & 20.1M & 10.48G\\
4 & \bf{7.50} & \bf{5.69} & \bf{0.38} & \bf{0.35} & 22.1M & 11.29G\\
\hline
\end{tabular}

}
\end{table*}

\begin{table*}[htb]
\centering

\caption{\label{tabDepSota} Performance of proposed and state-of-the-art methods for estimating of depression scores on AVEC datasets.}
\label{tabDep}
{
\begin{tabular}{l||cc|cc|c}
\hline
\multirow{2}{*}{\bf{Architecture}} & \multicolumn{2}{c|}{\textbf{AVEC2013}} & \multicolumn{2}{c|}{\textbf{AVEC2014}} & \multirow{2}{*}{\bf{Parameters $\downarrow$}} \\\cline{2-5}
& \bf{RMSE} & \bf{MAE} & \bf{RMSE} & \bf{MAE} & \\

\hline
\hline

Baseline-AVEC2013~\cite{avec2013} & 13.61 & 10.88 & - & - & - \\
Baseline-AVEC2014~\cite{avec2014} & - & - & 10.86 & 8.86 & - \\
MHH + LBP~\cite{Meng} & 11.19 & 9.14 & - & - & -\\
LPQ + Geo + CCA~\cite{Kaya}       & 9.72 & 7.86 & - & -  & - \\
Two-stream GoogLeNet~\cite{twoStreamDep}    & 9.82 & 7.58 & 9.55 & 7.47 & -\\
Two C3D~\cite{c3dDep} & 9.28 & 7.37 & 9.20 & 7.22 & $\approx$64.2M \\
Two C3D~\cite{fg} & 8.26 & 6.40 & 8.31 & 6.59 & $\approx$64.2M \\
VGG-16 + FDHH~\cite{FDHHDep} & - & - & 8.04 & 6.68 & $\approx$138.0M \\
DTL~\cite{DTLDep} & - & - & 9.43 & 7.74 & - \\
ResNet-50 + pooling~\cite{PoolingDep} & - & - & 8.43 & 6.37 & $\approx$23.5M  \\
Four ResNet-50~\cite{FourResNetDep} & 8.28 & 6.20 & 8.39 & 6.21 & $\approx$94.0M  \\
ResNet-50~\cite{icip} & 8.25 & 6.30 & 8.23 & 6.15 & $\approx$23.5M \\
Behavior signals~\cite{song} & 8.10 & 6.16 & 8.30* & 6.78* & - \\
DLGA-CNN~\cite{attentionDep} & 8.39 & 6.59 & 8.30 & 6.51 & - \\
Two-stream ResNet-50~\cite{icassp} & 7.97 & \bf{5.96} & 7.94 & 6.20 & $\approx$47.0M \\
MSN~\cite{msn} & 7.90 & 5.98 & 7.61 & 5.82& $\approx$77.7M \\
MDN~\cite{mdn} & \bf{7.55} & 6.24 & 7.65 & 6.06 & $\approx$52M\\
\hline
DMSN (Ours) & 7.66 & 6.14 & \bf{7.50} & \bf{5.69} & 22.1M\\
\hline
\multicolumn{6}{l}{* \footnotesize Results of the method for Freeform task.}
\end{tabular}
}
\end{table*}

\subsection{Analysis of multiscale spatiotemporal ability}
Given that facial dynamics comprise different spatiotemporal variations, it is essential that our architecture has a multiscale spatiotemporal representation ability to encode such variations. We evaluate this ability in our architecture by changing the number of branches in the Main Stage sub-block. To maintain a comparable computational complexity, when the number of branches is reduced, we increase the number of channels in the branches of the Main Stage sub-block. As seen in Table~\ref{tab5}, for depression detection and pain estimation, when more spatiotemporal ranges are explored (i.e., increasing the number of branches), the performance of DMSN improves, indicating a boost in the ability of encoding facial variations. It is worth noting that by increasing the number of branches in the Main Stage sub-block, the architecture not only enhances the capacity of exploring spatiotemporal features in different ranges, but it also increases the diversity of this exploration, since DMSN employs DMSN-A, DMSN-B, and DMSN-C blocks, which use different strategies to learn multiscale spatiotemporal features.





\subsection{Comparison with state-of-the-art}
In this section, the performance of our DMSN architecture is compared with state-of-the-art methods for depression detection and pain estimation.

\subsubsection{Depression detection}
Table~\ref{tabDepSota} compares the performance of our proposed architecture with state-of-the-art methods on AVEC2013 and AVEC2014 depression datasets. DMSN outperforms the method based on LPQ features~\cite{Kaya} and other related descriptors~\cite{avec2013,avec2014,Meng}. The methods in~\cite{FDHHDep,DTLDep,PoolingDep,FourResNetDep,icip,attentionDep} are based on 2D CNNs followed by an aggregation technique. The methods in~\cite{c3dDep,fg} employ 3D CNNs to explore spatiotemporal information. The authors in~\cite{twoStreamDep,icassp} infer depressive states by using two-stream networks. Our DMSN outperforms these methods (except for the method in~\cite{icassp} in terms of MAE on AVEC2013, but this approach employs $24.9$M more parameters). These results confirm findings in~\cite{msn,mdn,song}, which underscore  the importance of a multiscale approach for facial depression recognition. We also observe that DMSN achieves better results than the method in~\cite{song}, which explores behavioral primitives (facial action units, head pose, and gaze directions). When compared with MSN~\cite{msn} and MDN~\cite{mdn}, DMSN outperforms both models on AVEC2014, and achieves competitive results on AVEC2013, while requiring $3.51\times$ and $2.35\times$ fewer parameters than MSN and MDN, respectively. These results show that our DMSN architecture can provide a cost-effective solution for depression detection.

\subsubsection{Pain estimation}

Table~\ref{tabPain1} compares the performance of our proposed architecture with state-of-the-art methods on UNBC-McMaster dataset. As can be seen, our DMSN outperforms different schemes for pain expression recognition. For instance, DMSN achieves better results than the method in~\cite{VGG16LstmPain}, which uses VGG-16 architecture and LSTM, while requiring around $6.2$ times fewer parameters. The comparison with the SCN method~\cite{scn} is interesting because the basic block of this architecture is composed of parallel 3D convolutions with diverse temporal depths to explore multiscale spatiotemporal information. DMSN presents similar performance as this method with significant reduction of parameters (DMSN has around $26.5\times$ fewer parameters). These results indicate that our DMSN architecture is also an efficient option for pain estimation.

\begin{table}[!h]
\centering
{

\caption{\label{tabPain1} Performance of our proposed DMSN architecture against state-of-the-art methods on UNBC-McMaster dataset.}

\begin{tabular}{l||cc|c}
\hline
\bf{Architecture} & \bf{MSE} & \bf{MAE} & \bf{Parameters $\downarrow$} \\
\hline
\hline
RVR+LBP+DCT~\cite{lbpPain1} & 1.39 & - & - \\
HoT~\cite{hotPain} & 1.21 & - & - \\
OSVR~\cite{gaborPain} & - & 0.81 & - \\
RCNN~\cite{RcnnPain} & 1.54 & - & - \\
VGG-11+LSTM~\cite{Vgg11LstmPain} & 1.22 & 0.58 & $\approx$133M \\
VGG-16+LSTM~\cite{VGG16LstmPain} & 0.74 & 0.5 & $\approx$138M \\
C3D~\cite{scn} & 0.71 & - & $\approx$32M \\
I3D~\cite{weaklyPain} & - & 0.80 & $\approx$13M \\
MDN~\cite{mdn} & 0.68 & 0.42 & $\approx$52M\\
SCN~\cite{scn} & \bf{0.32} & - & $\approx$586.8M \\
\hline
DMSN (Ours)  & 0.38 & \bf{0.35}        & 22.1M  \\

\hline
\end{tabular}

}
\end{table}

Table~\ref{tabPain2} compares our DMSN architecture with state-of-the-art methods on BioVid dataset. DMSN outperforms the method in~\cite{daPain} which also explores facial expressions variations from videos. In~\cite{fusionPain}, the authors explore diverse features from ECG, EMG, and SCL as well as face videos. As we can see, DMSN obtains comparable results, demonstrating that facial expression analysis can provide essential information for the estimation of pain intensities.

\begin{table}[!h]
\centering
{

\caption{\label{tabPain2} Performance of our proposed DMSN architecture against state-of-the-art methods on BioVid dataset.}

\begin{tabular}{l|c||cc}
\hline
\bf{Method} & \bf{Modality} &\bf{MSE} & \bf{MAE} \\
\hline
\hline
I3D~\cite{daPain} & Video & - & 1.42 \\
Fusion~\cite{fusionPain} & Multimodal & 1.16 (RMSE)  & \bf{0.99}\\
\hline
DMSN (Ours) & Video & 1.54 & 1.04\\
\hline
\end{tabular}

}
\end{table}

\begin{figure*}[htb]
\centering
\includegraphics[scale=0.32]{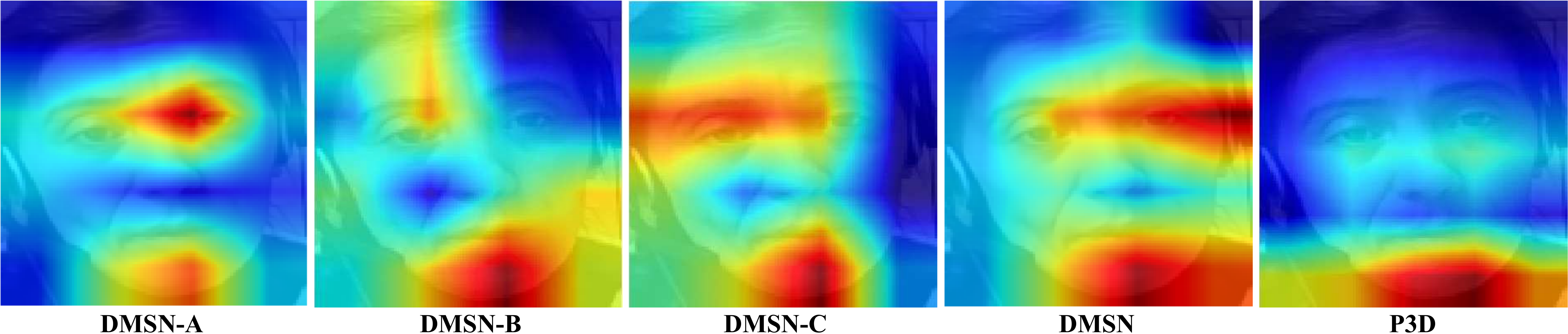}
\caption{Example of the CAMs showing the facial regions activated by our proposed DMSN models and P3D on a facial image from the AVEC2014 dataset.}
\label{figDep}
\end{figure*}

\begin{figure*}[htb]
\centering
\includegraphics[scale=0.281]{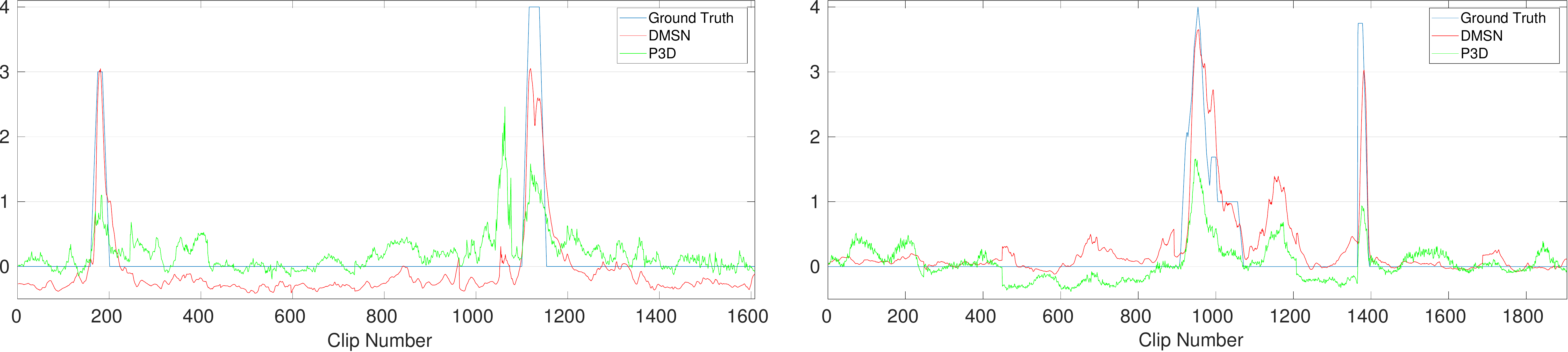}
\caption{Performance analysis of our proposed DMSN architecture and P3D on samples of two individuals from UNBC-McMaster dataset.}
\label{figPain}
\end{figure*}

\subsection{Cross-database analysis}
In order to assess the generalization capabilities of our DMSN architecture, we perform cross-database experiments. In this procedure, the source and target databases can belong to different tasks (e.g., AVEC2013 is the source database, and UNBC-McMaster is the target database). In this case, since the labels of pain and depression datasets are different, we replace the regression layer of DMSN to properly evaluate the representations generated by the model. In Table~\ref{tabCross}, we present the results of this experiment. When the evaluations are performed in the same task (e.g., depression detection), the model achieves reasonable results, indicating a robust representation for facial videos. The analysis between tasks is interesting because it allows an investigation about the applicability of depression/pain features to the pain/depression recognition task. We can observe that the representations learned on depression datasets allow DMSN to achieve good results on pain datasets. On the other hand, when DMSN is trained on pain datasets and then evaluated on depression detection task,  there is a higher degradation in performance. One reason for this result is the high level of ambiguity in depressive states which makes it difficult to directly apply the features of other applications.

\begin{table}[htb]
\centering
\caption{\label{tabCross} Performance of the proposed method in cross-dataset setting.}
{
\begin{tabular}{cc|ccc}
\hline
\bf{Training set} & \bf{Test set} &  \bf{RMSE}  &  \bf{MAE} & \bf{MSE}\\\hline
\hline
AVEC2013 & AVEC2014 & 7.78 & 6.18 & -\\
AVEC2014  & AVEC2013 & 8.36 & 6.62 & -\\
UNBC & BioVid & - & 1.19 & 1.92\\
BioVid & UNBC & - & 0.63 & 0.91\\
AVEC2013  & UNBC & - & 0.62 & 0.92\\
AVEC2014  & UNBC & - & 0.61 & 0.90\\
AVEC2013  & BioVid & - & 1.19 & 1.95\\
AVEC2014  & BioVid & - & 1.21 & 1.99\\
UNBC & AVEC2013 & 11.13 & 9.41 & -\\
UNBC & AVEC2014 & 11.24 & 9.40 & -\\
BioVid & AVEC2013 & 11.10 & 9.27 & -\\
BioVid & AVEC2014 & 10.93 & 9.13 & -\\\hline

\end{tabular}
}
\end{table}

\subsection{Qualitative results}

To interpret the performance differences for depression detection between our DMSN architecture and DMSN-A, DMSN-B, DMSN-C models as well as P3D, we present the class activation maps (CAMs) employing the Grad-CAM method~\cite{cam}. In the visualizations of Fig.~\ref{figDep}, lighter colors represent those regions that are most relevant for a model's predictions. Considering the most activated regions, the models appear to explore the eyes and mouth regions. In fact, these regions convey important information about depressive states. As we can see, our approach is more effective in exploring such areas than
P3D. In comparison with DMSN-A, DMSN-B, and DMSN-C, DMSN seems to be more successful in capturing face expression variations from these areas. We understand that this capacity of DMSN is a decisive factor for the good performance in depression detection.

Fig.~\ref{figPain} shows the effectiveness of our approach for pain estimation by comparing the predictions of our architecture with that of P3D and ground truth. It can be observed that our architecture can satisfactorily identify the occurrence of pain, which is an important characteristic for clinical application, whereas P3D has lower accuracy. DMSN presents a better
performance than P3D in recognizing changes of pain levels, which is due to a better multiscale spatiotemporal ability of DMSN. In general, our architecture has a good ability to follow the variations of pain levels, meaning that DMSN is effectively modeling transitions in facial pain expressions.

\section{Conclusion} \label{sec:con}
In this paper, we propose a structure called Decomposed Multiscale Spatiotemporal Network (DMSN) to learn multiscale spatiotemporal features from facial expressions in videos. Three variants of the DMSN block are introduced, which employ different strategies to effectively and efficiently capture facial dynamics. We design our DMSN architecture using these blocks to explore a variety of multiscale spatiotemporal features, which favors the adaptation to different facial behaviors. In our extensive experiments on AVEC2013 and AVEC2014 depression datasets, and UNBC-McMaster and BioVid pain datasets, we show that exploring the spatiotemporal information at multiple spatial sizes (DMSN-C block) is effective for depression detection, whereas capturing spatiotemporal features at multiple temporal ranges (DMSN-A block) is efficient for pain estimation. We also show that our architecture achieves competitive performance against state-of-the-art approaches for depression and pain expression detection, yet requires significantly fewer model parameters. Moreover, we demonstrate that depression features are more useful for pain estimation than pain features are for depression detection. In future work, we plan to investigate the performance of our DMSN architecture in other healthcare applications such as stress detection.

\bibliographystyle{elsarticle-num}

\bibliography{cas-refs}



\end{document}